\documentclass[sigconf,nonacm]{acmart}

\AtBeginDocument{%
  }

\renewcommand\footnotetextcopyrightpermission[1]{}

\usepackage{multirow}%

\begin{document}

\title{SKG-VLA: Scene Knowledge Graph Priors for Structured Scene Semantics and Multimodal Reasoning for Decision Making}

\author{Zeyu Li}
\affiliation{%
  \institution{Beijing University of Posts and Telecommunications}
  \city{Beijing}
  \country{China}
}
\email{lizeyu@bupt.edu.cn}

\author{Lei Li}
\affiliation{%
  \institution{Beijing University of Posts and Telecommunications}
  \city{Beijing}
  \country{China}
}
\email{leili@bupt.edu.cn}

\renewcommand{\shortauthors}{Li and Li}

\begin{abstract}
Decision making in large-scale complaint handling systems increasingly relies on heterogeneous evidence, including complaint narratives, screenshots, order metadata, historical interactions, and platform policies. Existing complaint understanding systems mainly perform shallow classification or template matching over isolated modalities, while underutilizing explicit scene structure, rule knowledge, and cross-evidence dependencies. To address this limitation, we present SKG-VLA for multimodal complaint decision making. The core idea is to model each case as a structured complaint scene and represent its decision-relevant semantics with a \emph{Scene Knowledge Graph} (SKG), which organizes complaint entities, evidence items, policy clauses, temporal events, transactional states, and action-relevant relations into a unified graph. Based on SKG, we build a data synthesis pipeline that generates complaint scene descriptions, rule-consistent graph generalizations, question--answer supervision, and decision recommendations. We further construct a large-scale complaint scene dataset with both text-only and multimodal in-domain benchmarks. Finally, we adopt a three-stage training strategy---domain-adaptive pre-training, task-oriented instruction fine-tuning, and end-to-end multimodal alignment---to inject structured scene priors into a multimodal decision model. Experiments show that SKG-VLA consistently improves policy-grounded reasoning, complaint decision accuracy, long-tail generalization, and robustness under incomplete evidence.

\end{abstract}



\maketitle
\begingroup
\renewcommand\thefootnote{}\footnotetext{Accepted by ICMR 2026. This is the author version for arXiv.}
\endgroup

\section{Introduction}

Modern online service platforms receive large volumes of user complaints involving products, logistics, payments, after-sales service, billing disputes, fraud suspicion, and content moderation disputes. In real applications, a complaint case is rarely represented by plain text alone. Instead, decision making often relies on heterogeneous evidence, such as complaint narratives, screenshots, order records, timestamps, previous negotiation history, and policy clauses. The final action must answer not only \emph{what happened}, but also \emph{what should be done}: whether the case should be refunded, compensated, transferred, escalated, rejected, or sent for manual review.
\begin{figure}[t]
  \centering
  \includegraphics[width=\linewidth]{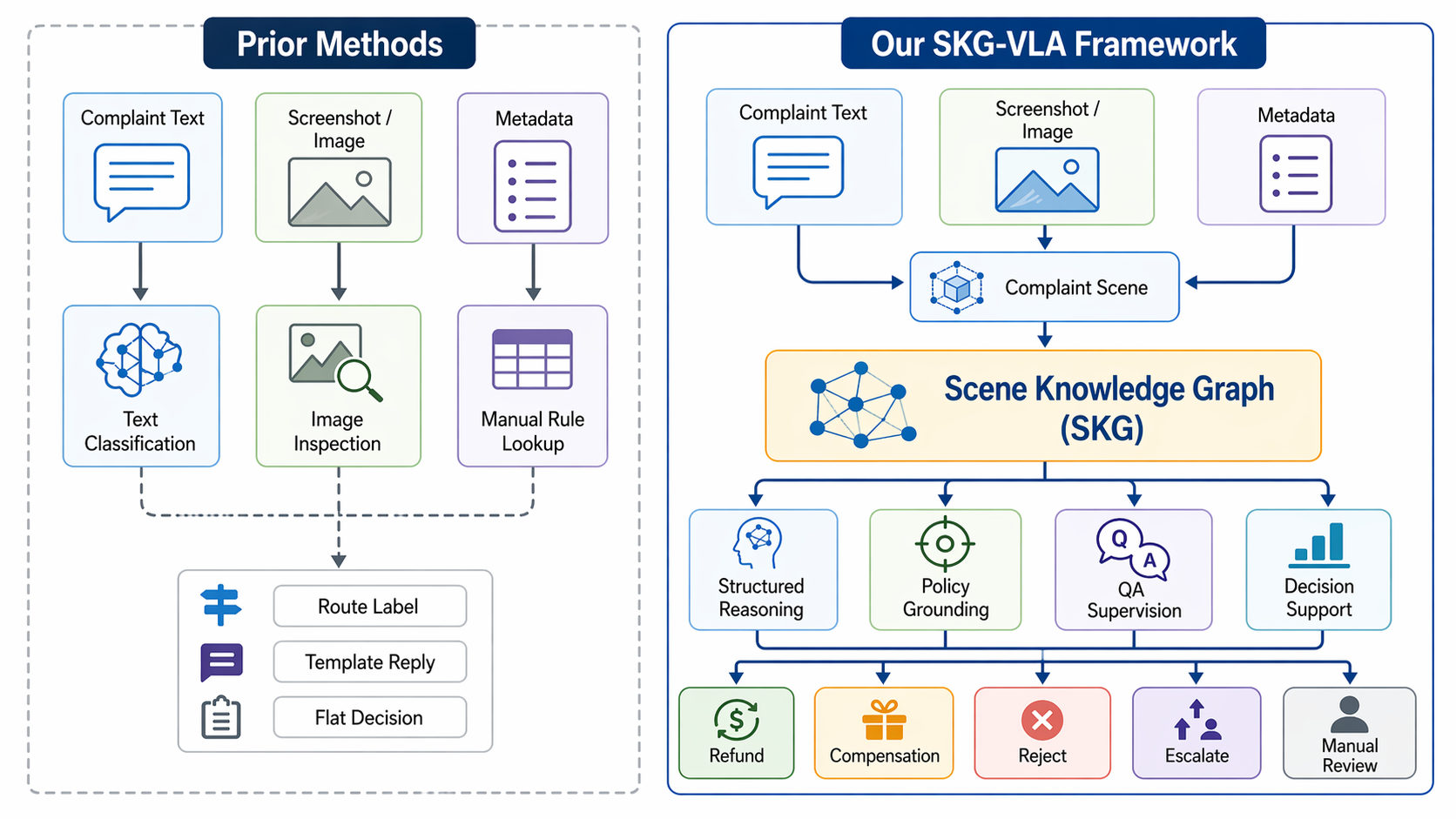}
  \caption{Comparison between prior multimodal complaint-processing pipelines and the proposed SKG-VLA framework. Existing systems often process complaint narratives, screenshots, and metadata independently and then map them to flat routing labels or template responses. In contrast, our method models each case as a complaint scene and represents its structured scene semantics with a Scene Knowledge Graph, enabling policy grounding, structured reasoning, and interpretable action recommendation.}
  \Description{A single-column opening figure comparing prior complaint processing methods with the proposed scene knowledge graph based framework.}
  \label{fig:comparison}
\end{figure}
Despite recent progress in multimodal large models, complaint decision making remains difficult for two reasons. First, existing systems still emphasize loosely structured supervision and surface-level matching between complaint text and response templates. They underexploit explicit scene semantics, policy knowledge, temporal relations, and cross-evidence dependencies. Second, complaint cases are inherently structured and rule-constrained. The validity of a decision depends on coordinated reasoning over complaint category, evidence sufficiency, service stage, responsibility attribution, user history, and applicable rules. Without explicit structure, a model may produce plausible-sounding yet policy-inconsistent recommendations.
\begin{figure*}[t]
  \centering
  \includegraphics[width=\textwidth]{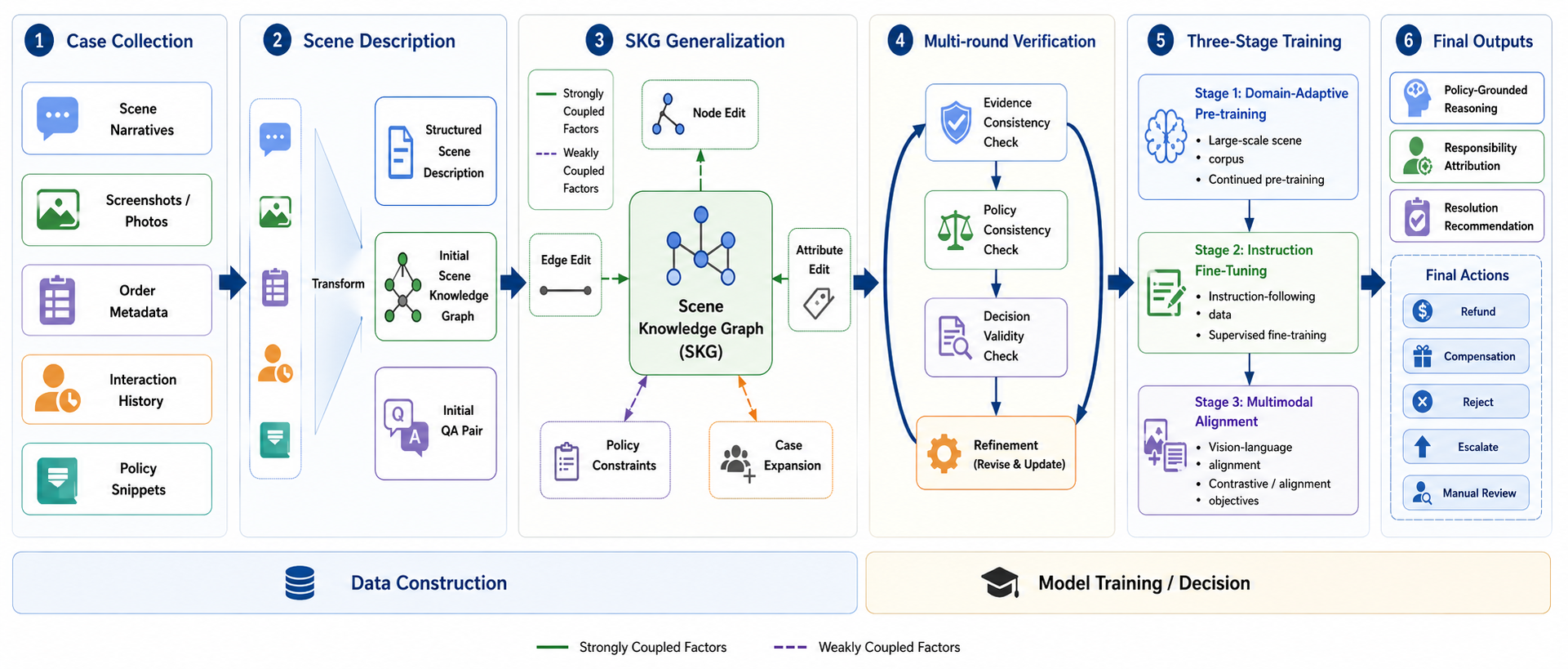}
  \caption{Overview of the SKG-VLA framework for multimodal complaint decision making. Starting from multimodal complaint evidence, the framework first constructs structured complaint scene descriptions and Scene Knowledge Graphs, then performs rule-consistent graph generalization and multi-round verification, and finally injects structured scene priors into the model through domain-adaptive pre-training, task-oriented instruction tuning, and end-to-end multimodal alignment.}
  \Description{A double-column pipeline figure showing the overall SKG-VLA framework from complaint evidence collection to scene graph construction, generalization, verification, and three-stage model training.}
  \label{fig:pipeline}
\end{figure*}

In this paper, we view each case as a \emph{complaint scene}: a structured operational situation jointly defined by user claims, visual evidence, transactional context, temporal evolution, policy state, and candidate actions. Under this view, complaint handling is no longer simple classification, but multimodal reasoning over structured scene semantics. This motivates \emph{Scene Knowledge Graphs} (SKGs), which explicitly organize complaint entities, event states, evidence items, policy constraints, and decision-relevant relations into a unified semantic graph. As illustrated in Fig.~\ref{fig:comparison}, the central difference from prior pipelines is that SKG-VLA does not treat complaint narratives, screenshots, and metadata as isolated evidence streams, but instead integrates them into a structured scene representation for downstream reasoning and action recommendation.

Public complaint and customer-service resources provide important building blocks but do not directly solve this problem. The CFPB Consumer Complaint Database offers large-scale real complaint narratives and company response fields \cite{cfpb-complaints}. JDDC, JDDC 2.0, MMD, SIMMC 2.0, and MultiWOZ provide realistic customer-service or task-oriented interaction data \cite{jddc,jddc2,mmd,simmc2,multiwoz}. However, these resources are not designed around multimodal complaint decision making with explicit scene structure, policy constraints, and auditable action supervision.

To address this gap, we propose SKG-VLA, a structured-prior framework for multimodal complaint decision making. The key representation in our method is a Scene Knowledge Graph that captures the structured scene semantics of each complaint case. Starting from multimodal evidence, the framework constructs complaint scene descriptions, scene graphs, rule-consistent scene variants, question--answer supervision, and final decision targets for multimodal learning. The overall pipeline is summarized in Fig.~\ref{fig:pipeline}, which links complaint evidence collection, SKG construction, graph generalization, verification, and three-stage training within a unified framework.

The contributions of this paper are summarized as follows:
\begin{itemize}
\item We reformulate multimodal complaint handling as a structured scene reasoning problem and introduce \emph{Scene Knowledge Graphs} as explicit priors for complaint scene understanding, policy grounding, and action recommendation.
\item We present a data synthesis framework that generates complaint scene descriptions, rule-consistent graph generalizations, QA supervision, and decision recommendations from multimodal complaint evidence.
\item We construct a large-scale complaint scene dataset and two in-domain evaluation suites, including a text-only benchmark and a multimodal complaint decision benchmark.
\item We design a three-stage training pipeline that progressively transfers structured scene priors into a multimodal decision model, improving interpretability, robustness, and policy consistency.
\end{itemize}

\section{Related Work}

\subsection{Complaint Understanding, Service Mining, and Decision Support}

Complaint understanding has traditionally been studied through issue classification, sentiment analysis, ticket routing, and anomaly mining. Early work mainly focused on textual complaint narratives and imbalanced label distributions \cite{tang2021complaint,gao2023anomaly}. Public resources such as the CFPB Consumer Complaint Database provide large-scale real-world complaint narratives and operational fields that support complaint analytics, trend analysis, and issue prediction \cite{cfpb-complaints}. More recently, LLM-based studies show that reasoning-oriented prompting and stronger language backbones improve complaint detection and classification, especially in policy-sensitive domains \cite{roumeliotis2025complaint}. Nevertheless, these studies remain largely text-only and typically stop at classification rather than full complaint decision support.

A closely related line of work studies customer-service and task-oriented dialogue. Datasets such as MultiWOZ, JDDC, JDDC 2.0, MMD, and SIMMC 2.0 have enabled research on dialogue state tracking, multimodal customer assistance, and product-grounded interaction \cite{multiwoz,jddc,jddc2,mmd,simmc2}. These resources are important for modeling realistic service language and multimodal context, but they are not designed around complaint-specific policy grounding, evidence sufficiency, responsibility attribution, or auditable action recommendation.

\subsection{Multimodal Evidence Understanding and Document-Centric Models}

Complaint handling often depends on screenshots, receipts, chat records, and UI evidence, which makes document and text-rich visual understanding highly relevant. Earlier benchmarks such as TextVQA, ST-VQA, OCR-VQA, and DocVQA emphasize reading and reasoning over scene text or document images \cite{textvqa,stvqa,ocrvqa,docvqa}. Document-centered foundation models further improve OCR-aware or OCR-free understanding, including LayoutLMv3, Donut, and Pix2Struct \cite{layoutlmv3,donut,pix2struct}. More recent work pushes this direction toward stronger end-to-end models and multi-page understanding, such as GOT-OCR 2.0 and mPLUG-DocOwl2 \cite{gotocr2,docowl2}. These advances are highly relevant for screenshot-style complaint evidence, yet they do not directly address the structured decision logic required by complaint handling.

\subsection{Multimodal Large Models and Reasoning Benchmarks}

General-purpose multimodal large language models (MLLMs) have rapidly improved open-world perception and instruction following. Representative systems include Flamingo, BLIP-2, LLaVA, LLaVA-1.5, and Qwen-VL \cite{flamingo,blip2,llava,llava15,qwenvl}. More recent open models such as Qwen2-VL, Qwen2.5, Qwen2.5-VL, and Molmo substantially strengthen visual grounding, document understanding, and long-context multimodal reasoning \cite{qwen2vl,qwen25,qwen25vl,molmo}. In parallel, evaluation suites such as MME, MMMU, and MMMU-Pro make it easier to measure perception, cognition, and discipline-specific reasoning in modern MLLMs \cite{mme,mmmu,mmmupro}. These models and benchmarks demonstrate strong generic capability, but complaint decision making differs from broad instruction following because the target action must remain consistent with domain rules, temporal evidence, and responsibility constraints.

\subsection{Knowledge-Enhanced and Graph-Augmented Reasoning}

Structured knowledge has long been used to improve transparency and factual consistency in decision systems. Knowledge-enhanced pretraining methods such as ERNIE, K-BERT, and KEPLER inject entity-level or relation-level priors into language models \cite{ernie,kbert,kepler}. More recent LLM-centered approaches reason over structured inputs or external knowledge sources through retrieval or tool use, including Chain-of-Thought prompting, ReAct, Toolformer, RAG, Self-RAG, and StructGPT \cite{cot,react,toolformer,rag,selfrag,structgpt}. Graph-augmented reasoning has also become increasingly important in 2024--2025, with systems such as GraphRAG, HippoRAG, G-Retriever, and MindMap showing that graph-structured retrieval and knowledge-guided prompting can improve multi-hop reasoning, evidence integration, and interpretability \cite{graphrag,hipporag,gretriever,mindmap}. Our framework is aligned with this trend, but differs in two ways: it represents the full complaint scene rather than a generic knowledge store, and it injects structured scene priors into both data synthesis and multimodal training rather than using graph structure only at inference time.

\section{Significance of Structured Scene Semantics in Complaint Decision Making}

Although screenshots and transactional metadata provide important evidence, language remains central in complaint decision making. User intent is most directly expressed through complaint narratives and historical dialogue, while platform policies, eligibility conditions, exception rules, and escalation criteria are themselves written in natural language. However, language alone is not sufficient: a complaint case typically involves multiple interdependent factors, including order state, delivery promise, service history, product condition, evidence completeness, timeline consistency, and policy applicability.

\begin{figure}[t]
  \centering
  \includegraphics[width=\linewidth]{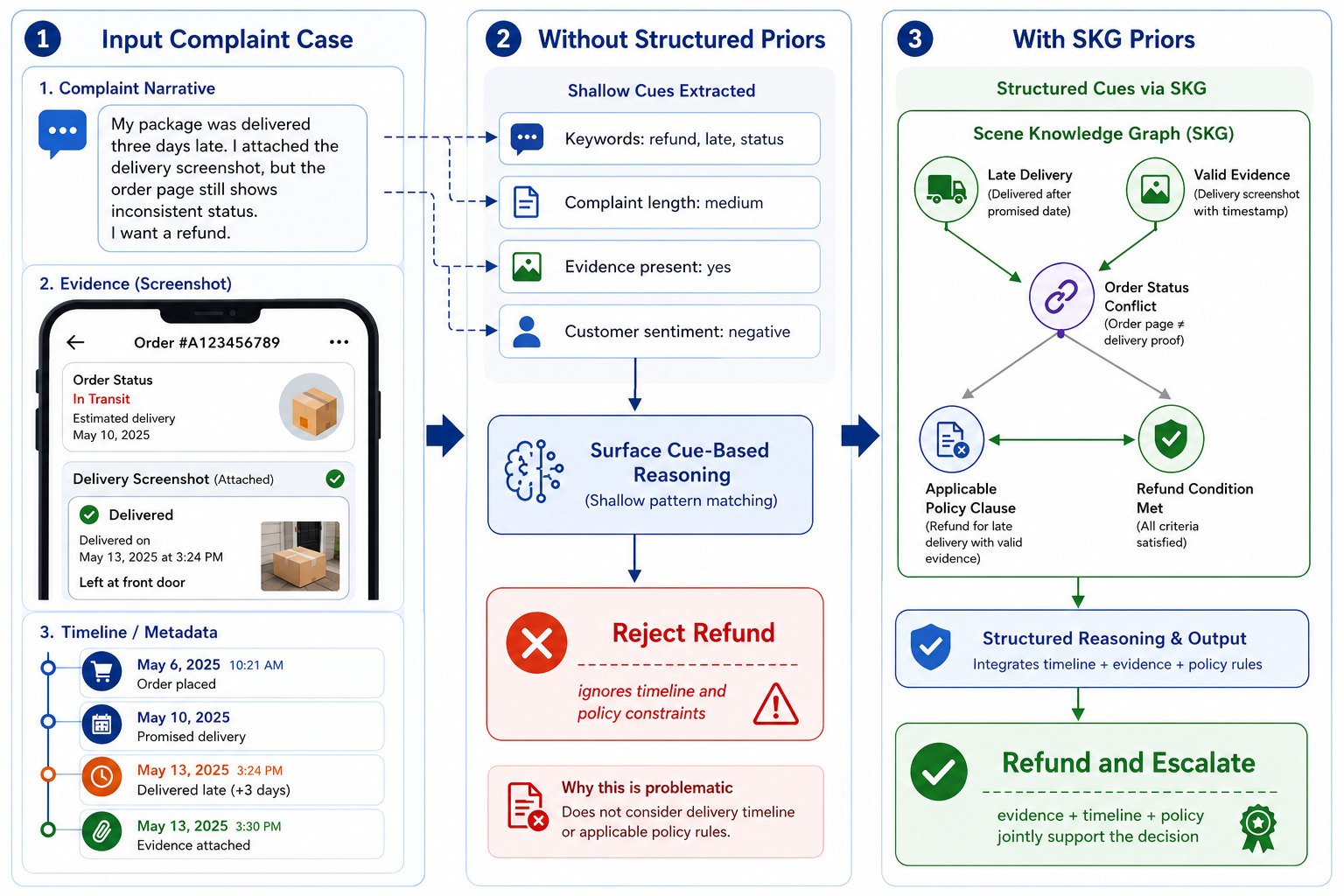}
  \caption{Illustration of the significance of structured scene semantics in complaint decision making. A baseline without explicit scene structure may rely on superficial cues and produce policy-inconsistent decisions, whereas the SKG-enhanced model reasons over complaint narratives, timeline evidence, and policy clauses to reach a more reliable and explainable action recommendation.}
  \Description{A single-column qualitative figure showing why structured scene semantics are important for complaint decision making.}
  \label{fig:significance}
\end{figure}

These observations motivate the introduction of Scene Knowledge Graphs as an intermediate representation for complaint decision making. In our framework, an SKG connects complaint entities, evidence items, temporal events, policy clauses, transactional states, and decision constraints into explicit scene-level relations. This structured representation enables policy-grounded reasoning, interpretable decision support, and rule-consistent case generalization more effectively than shallow textual matching. Fig.~\ref{fig:significance} qualitatively illustrates how structured scene semantics can alter both the reasoning path and the final action outcome compared with a baseline that relies mainly on superficial cues.

\section{Dataset and Benchmarks}

\subsection{Public-Source Pool and Construction Protocol}

To build a complaint-domain dataset that is rich in evidence and explicit in decision logic, we combine a public-source pool with an internal scene-construction and verification pipeline. We draw seed complaint narratives and resolution-related fields from the CFPB Consumer Complaint Database \cite{cfpb-complaints}. We use JDDC, JDDC 2.0, MMD, and SIMMC 2.0 to diversify service language, multimodal interaction patterns, and grounded dialogue structure \cite{jddc,jddc2,mmd,simmc2}. For screenshot-style and document-like evidence understanding, we additionally reference document and text-rich visual QA resources such as DocVQA and TextVQA \cite{docvqa,textvqa}.

The annotation pipeline has three stages. First, an LLM-based parser produces a complaint scene description, candidate entities, timeline events, evidence tags, transactional states, policy clauses, and action candidates. Second, annotators verify the consistency between the source evidence and the generated scene graph. Third, senior reviewers audit the final graph, QA pair, and recommended action to ensure alignment with policy constraints and operational decision logic. To improve coverage beyond naturally occurring complaint patterns, we further perform structured graph generalization rather than plain textual paraphrasing.

\subsection{Dataset Statistics and Positioning}

The dataset used in this paper contains \textbf{183,642} verified base complaint cases, which are expanded into \textbf{1,428,916} Scene Knowledge Graphs and the same number of decision-oriented QA pairs. We further construct a text-only benchmark, \textbf{ComplaintScene-Text}, with \textbf{3,286} cases and \textbf{7,504} QA pairs, and a multimodal benchmark, \textbf{ComplaintScene-MM}, with \textbf{6,914} multimodal cases and \textbf{18,237} QA pairs. Tab.~\ref{tab:dataset_comp} compares our dataset with representative public complaint and customer-service resources and highlights that our setting jointly covers multimodal evidence, complaint narratives, explicit policy signals, and action-level supervision.

\begin{table*}[t]
  \caption{Comparison with public complaint and customer-service resources. Public sizes are reported using the units used by the original sources.}
  \centering
  \resizebox{\textwidth}{!}{%
  \begin{tabular}{lcccccc}
    \toprule
    Dataset & Domain & Public Size / Unit & Image Evidence & Complaint Narratives & Explicit Policy Fields & Action-Level Labels \\
    \midrule
    CFPB Consumer Complaint Database \cite{cfpb-complaints} & Consumer finance complaints & Public, daily-updated, millions of records & No & Yes & Partial & Partial \\
    JDDC \cite{jddc} & E-commerce customer service & $>$1M dialogues, 20M utterances & No & Partial & No & No \\
    JDDC 2.0 \cite{jddc2} & Multimodal customer service & 246K sessions, 3M utterances, 507K images & Yes & Partial & Product KB only & No \\
    MMD \cite{mmd} & Retail multimodal dialogue & $>$150K sessions, $>$6.5M utterances & Yes & No & No & No \\
    SIMMC 2.0 \cite{simmc2} & Shopping multimodal dialogue & 11K dialogs, 117K utterances & Yes & No & No & No \\
    \midrule
    Ours (SKG dataset) & Complaint scene decision making & 183,642 cases, 1.43M SKGs, 1.43M QA pairs & Yes & Yes & Yes & Yes \\
    \bottomrule
  \end{tabular}}
  \label{tab:dataset_comp}
\end{table*}

\begin{figure}[t]
  \centering
  \includegraphics[width=\linewidth]{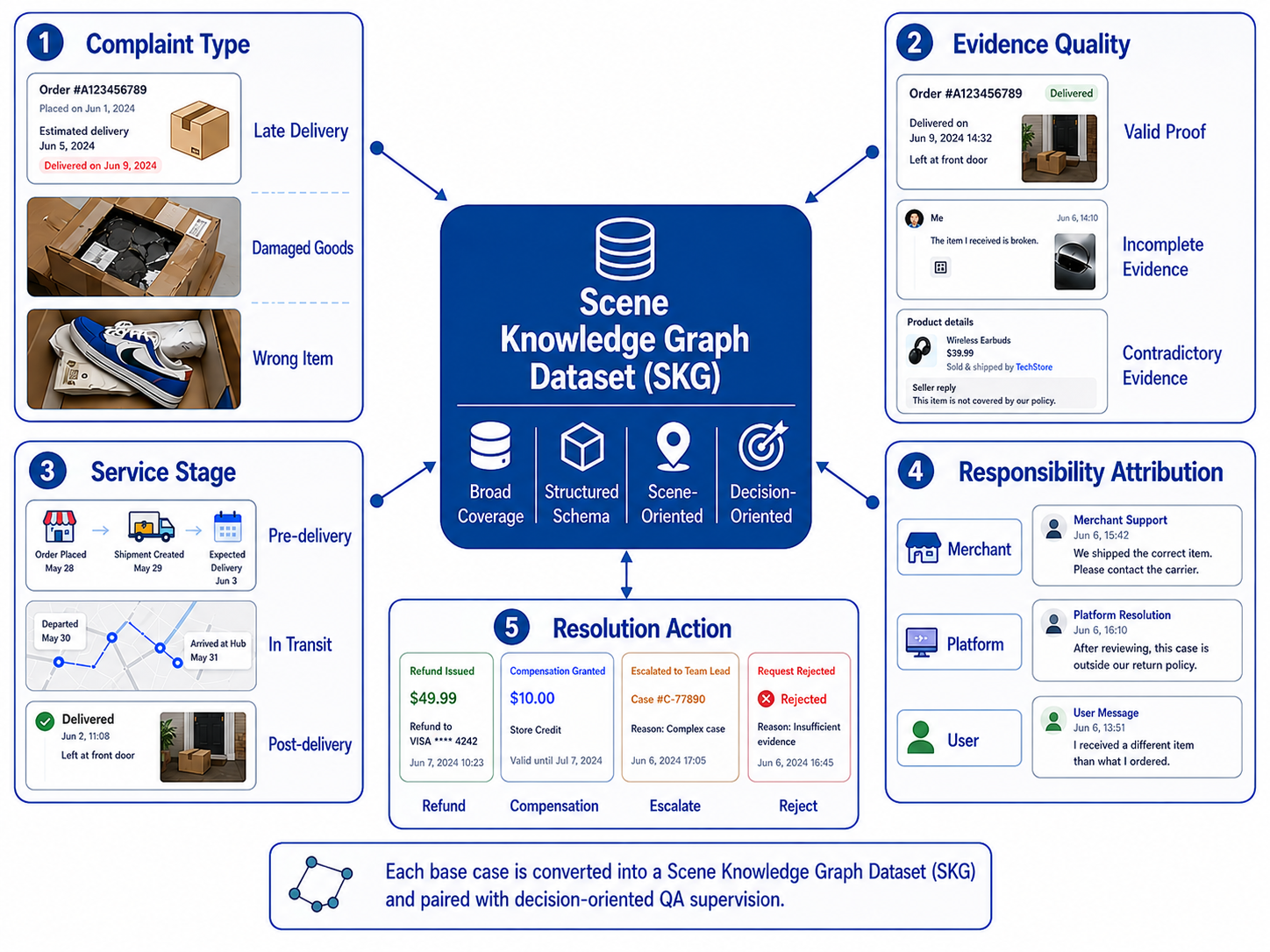}
  \caption{Overview of the proposed complaint scene dataset. Cases are organized by five decision-critical scene dimensions: complaint type, evidence quality, service stage, responsibility attribution, and resolution action. Each verified base case is converted into a canonical Scene Knowledge Graph and can then be generalized into multiple rule-consistent graph variants paired with decision-oriented QA supervision.}
  \Description{A single-column dataset overview figure for the complaint scene dataset.}
  \label{fig:dataset}
\end{figure}

Compared with earlier public resources, our dataset combines multimodal evidence, complaint narratives, explicit policy signals, structured scene graphs, and action-level decision supervision in a single framework. This is critical because complaint decision making depends not only on category recognition, but also on evidence sufficiency, responsibility attribution, and policy-consistent action selection. Fig.~\ref{fig:dataset} further illustrates how the dataset spans five decision-critical scene dimensions.

\subsection{In-Domain Benchmarks}

We define two in-domain evaluation suites.

\noindent\textbf{ComplaintScene-Text.}
This benchmark rewrites each multimodal complaint case into a structured complaint scene description that preserves complaint type, evidence status, timeline, transactional state, and policy cues. It presents multiple-choice or binary questions targeting evidence sufficiency, responsibility judgment, policy applicability, escalation need, and final action recommendation.

\begin{figure}[t]
  \centering
  \includegraphics[width=\linewidth]{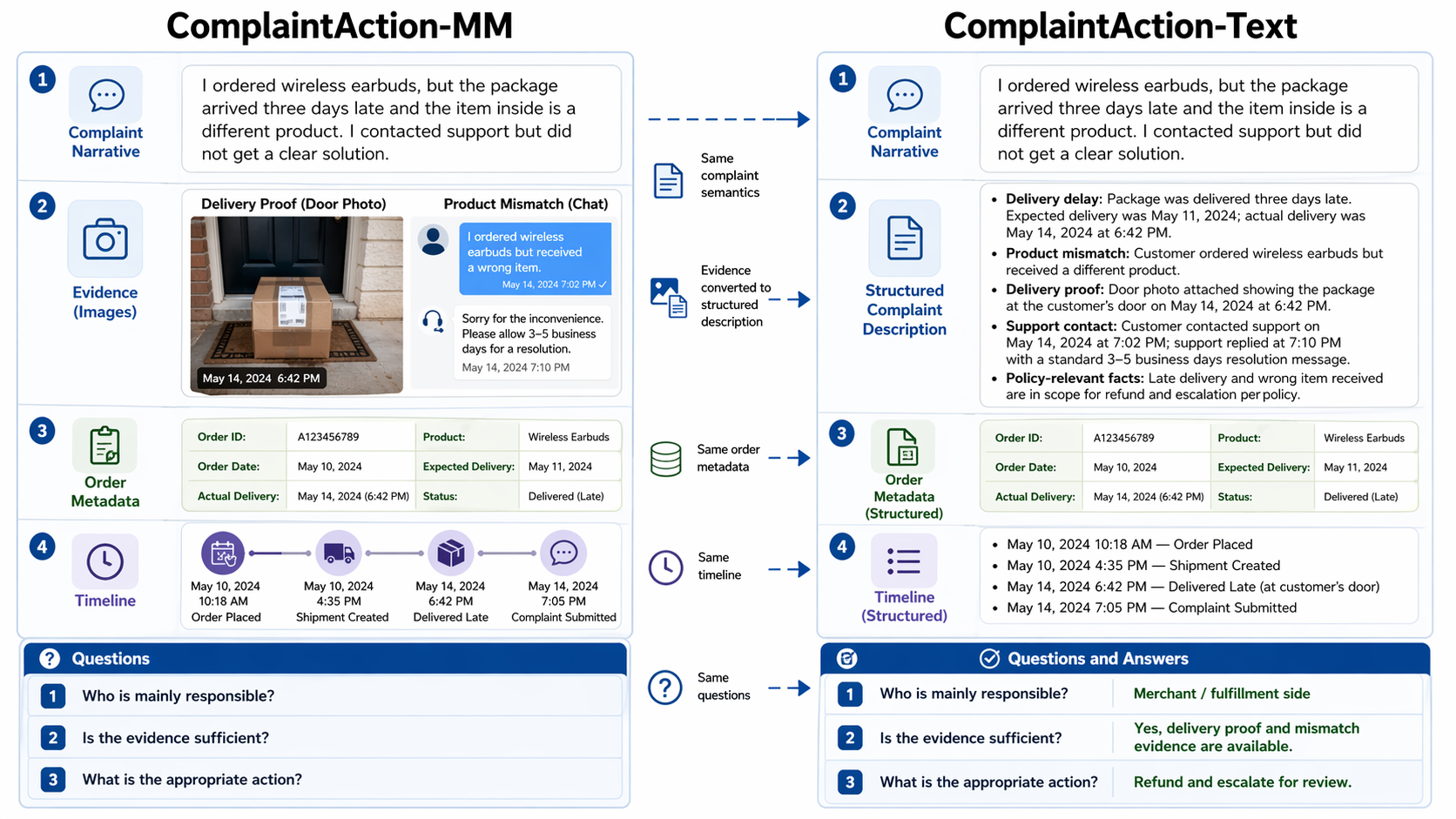}
  \caption{Comparison between the original multimodal complaint scene and the text-only benchmark version. The text-only benchmark preserves complaint scene semantics, key evidence, timeline structure, transactional state, and policy cues while removing raw visual evidence, allowing direct evaluation of language-grounded decision reasoning.}
  \Description{A figure comparing a multimodal complaint scene and its text-only benchmark version.}
  \label{fig:benchmark}
\end{figure}

\noindent\textbf{ComplaintScene-MM.}
This benchmark retains the original multimodal inputs, including complaint narratives, screenshots or photo evidence, metadata, and historical interaction summaries, and evaluates full complaint scene reasoning and decision capability under the realistic evidence-rich setting. Fig.~\ref{fig:benchmark} shows a paired example of the same complaint case under the multimodal and text-only settings, illustrating how scene semantics and policy cues are preserved after converting raw evidence into structured textual form.

\section{Method}

\subsection{Structured Data Generation}

Let a complaint case be denoted by
\begin{equation}
\mathbf{X}=\{x^{\text{text}},x^{\text{img}},x^{\text{meta}},x^{\text{hist}},x^{\text{rule}}\},
\end{equation}
where $x^{\text{text}}$ is the complaint narrative, $x^{\text{img}}$ is screenshot or photo evidence, $x^{\text{meta}}$ contains transactional and timeline metadata, $x^{\text{hist}}$ denotes historical interactions, and $x^{\text{rule}}$ denotes relevant policy clauses.

We interpret $\mathbf{X}$ as a \emph{complaint scene}, namely a structured operational situation jointly defined by user claims, visual evidence, transactional context, temporal progression, and platform rules. We represent this complaint scene with a Scene Knowledge Graph
\begin{equation}
G=(\mathcal{V},\mathcal{E},\mathcal{A}),
\end{equation}
where $\mathcal{V}$ denotes scene entities, evidence nodes, policy nodes, event nodes, state nodes, and decision nodes; $\mathcal{E}$ denotes typed relations; and $\mathcal{A}$ denotes node attributes.

Not all graph variables are equally important for the final decision. We therefore partition nodes into \emph{strongly coupled} and \emph{weakly coupled} categories according to decision sensitivity. Strongly coupled nodes include complaint type, order status, timeline events, evidence validity, policy clauses, service stage, and responsibility attribution. These nodes directly affect the direction and admissibility of the final decision and must be edited coordinately during graph generalization. Weakly coupled nodes include peripheral visual details, optional metadata fields, and stylistic wording variations.

Given multimodal case input $\mathbf{X}$ and a prompt template $P$, an initial generator produces a scene graph $G_0$, a structured scene description $D_0$, and a decision-oriented QA item $Q_0$:
\begin{equation}
(G_0,D_0,Q_0)=\mathrm{LLM}_{\mathrm{gen}}(\mathbf{X},P).
\end{equation}
To guarantee consistency between graph, evidence, and policy, we introduce an iterative verification loop:
\begin{equation}
V_k=\mathrm{LLM}_{\mathrm{verify}}(\mathbf{X},G_k,D_k,Q_k),
\end{equation}
\begin{equation}
(G_{k+1},D_{k+1},Q_{k+1})=\mathrm{LLM}_{\mathrm{refine}}(\mathbf{X},G_k,D_k,Q_k,V_k).
\end{equation}
The process runs for $K$ iterations and yields final outputs $(G_f,D_f,Q_f)$. This data construction process corresponds to the left half of Fig.~\ref{fig:pipeline}, where raw complaint evidence is converted into structured scene descriptions and SKGs and is then expanded through rule-consistent graph generalization and multi-round verification.

After verification, we expand the complaint scene space through rule-constrained graph edits:
\begin{equation}
\tilde{G}=\Phi(G_f;\mathcal{C}),
\end{equation}
where $\mathcal{C}$ is a set of policy constraints. This operation modifies evidence sufficiency, service stage, merchant response, event relations, or state variables while preserving logical validity. Each generalized graph is then converted into a new scene description, QA instance, and decision recommendation.

\subsection{Three-Stage Training}

To inject structured scene priors into a multimodal decision system, we adopt a three-stage training pipeline. As illustrated in Fig.~\ref{fig:pipeline}, this training process progressively transfers structured scene priors from language-only supervision to end-to-end multimodal decision learning.

\noindent\textbf{Domain-Adaptive Pre-training.}
We first adapt the language model to complaint-domain semantics using the generated scene descriptions and QA data:
\begin{equation}
\mathcal{L}_{\text{PT}}=-\sum_t \log P(w_t \mid w_{<t},D_f,Q_f;\Theta_{\text{LLM}}).
\end{equation}

\noindent\textbf{Task-Oriented Instruction Fine-Tuning.}
We then fine-tune the model to answer complaint-scene-related questions and output policy-grounded decisions:
\begin{equation}
\mathcal{L}_{\text{SFT}}=-\sum_t \log P(a_t \mid q,a_{<t},D_f,Q_f;\Theta_{\text{LLM}}).
\end{equation}

\noindent\textbf{End-to-End Multimodal Alignment.}
Finally, we connect the fine-tuned language model to a visual encoder and structured metadata projector:
\begin{equation}
\mathcal{L}_{\text{MM}}
=-\sum_t \log P(a_t \mid x^{\text{text}},x^{\text{img}},x^{\text{meta}},q;\Theta_{\text{enc}},\Theta_{\text{LLM}},\Theta_{\text{pro}}).
\end{equation}
This stage transfers complaint scene priors from the language space into a multimodal decision model capable of consuming raw evidence.

\section{Experiments}

\subsection{Experimental Setup}

We implement the SKG-VLA framework using an open-source multimodal training stack. We evaluate three text backbone scales, namely \textbf{Qwen2.5-0.5B}, \textbf{Qwen2.5-1.5B}, and \textbf{Qwen2.5-3B} \cite{qwen25}. Unless otherwise noted, all text-only models are initialized from the corresponding Qwen2.5 checkpoint. For multimodal experiments, the language backbone is initialized from the corresponding Qwen2.5 model, while the visual branch and multimodal connector follow the official Qwen2-VL / Qwen2.5-VL design \cite{qwen2vl,qwen25vl}. In addition to our own multimodal variants, we further compare against the official general-purpose open-source model \textbf{Qwen2.5-VL-3B-Instruct} as a strong external multimodal baseline at the 3B scale.

Accordingly, the entries labeled \textbf{llm}, \textbf{pre}, and \textbf{sft} in the result tables denote text-only models built on the same Qwen2.5 backbone, with increasing amounts of complaint-domain adaptation. Specifically, \textbf{llm} denotes the raw base model, \textbf{pre} denotes the model after domain-adaptive pre-training, and \textbf{sft} denotes the model after subsequent task-oriented instruction fine-tuning. The entries labeled \textbf{mm-ori} and \textbf{mm-our} denote multimodal models using the same language backbone at the corresponding scale. \textbf{mm-ori} is the multimodal baseline without our structured scene prior injection, while \textbf{mm-our} is the full SKG-VLA model with structured scene priors, graph-derived supervision, and complaint-specific multimodal adaptation.

For domain-adaptive language training, we adopt Megatron-LM style distributed training \cite{megatronlm}. During multimodal training, the visual encoder, projection layer, and language backbone are jointly optimized end to end. We use AdamW as the optimizer, BF16 mixed precision, and FlashAttention-2 for efficient attention computation \cite{adamw,flashattention2}.

All experiments are conducted on 2 nodes, each equipped with $8\times$ AMD Instinct MI308X GPUs (192GB HBM memory per GPU) and Intel Xeon Platinum 8575C processors. For domain-adaptive pre-training, we use a learning rate of $1\times10^{-4}$, sequence length 4096, and global batch size adjusted by model scale. For supervised instruction tuning, we use a lower learning rate of $5\times10^{-6}$ and sequence length up to 8192 to accommodate complaint narratives, policy snippets, and graph-derived scene descriptions. For multimodal alignment, sequence length and micro-batch size are tuned by model scale under the same hardware budget.

The dataset is split by base complaint case rather than by graph variant to avoid leakage between generalized cases derived from the same source case. Public-source derived CFPB tasks follow standard train/dev/test splits with company-aware and time-aware filtering. Our in-domain benchmarks are split by base complaint case into train/dev/test partitions.

\subsection{Evaluation Benchmarks and Metrics}

We evaluate the method from four perspectives.

\noindent\textbf{Public text-only complaint tasks.}
From the CFPB database, we construct two standard classification tasks: \textbf{CFPB-Product}, which predicts the official product field from the complaint narrative, and \textbf{CFPB-Issue}, which predicts the official issue field. We report accuracy for CFPB-Product and macro-F1 for CFPB-Issue.

\noindent\textbf{In-domain text-only benchmark.}
\textbf{ComplaintScene-Text (ours)} focuses on policy-grounded reasoning under a text-only complaint scene setting. On this benchmark, we evaluate three subtask dimensions: \emph{Evidence}, \emph{Policy}, and \emph{Action}. \emph{Evidence} measures the accuracy of evidence sufficiency judgment, \emph{Policy} measures the accuracy of policy applicability prediction, and \emph{Action} measures the macro-F1 of final action recommendation. Let $s_{\text{evi}}$, $s_{\text{pol}}$, and $s_{\text{act}}$ denote the corresponding scores. The reported text-only average is
\begin{equation}
\mathrm{Avg}_{\text{text}}=\frac{s_{\text{evi}}+s_{\text{pol}}+s_{\text{act}}}{3}.
\end{equation}
\begin{table*}[t]
  \caption{Main in-domain results with stepwise scene-prior injection and multimodal fusion. The text backbones are Qwen2.5-0.5B, Qwen2.5-1.5B, and Qwen2.5-3B. At the 3B scale, we additionally report Qwen2.5-VL-3B-Instruct as an official general-purpose multimodal baseline. Text-only columns are evaluated on ComplaintScene-Text, while multimodal columns are evaluated on ComplaintScene-MM.}
  \centering
  \resizebox{\textwidth}{!}{%
  \begin{tabular}{llcccc|cccc}
    \toprule
    \multicolumn{2}{c}{Model} & \multicolumn{4}{c}{Text-only (ComplaintScene-Text)} & \multicolumn{4}{c}{Multimodal (ComplaintScene-MM)} \\
    \cmidrule(lr){1-2}\cmidrule(lr){3-6}\cmidrule(lr){7-10}
    Backbone & Variant & Avg & Evidence & Policy & Action & Avg & Routing & Responsibility & Resolution \\
    \midrule
    \multirow{5}{*}{Qwen2.5-0.5B}
    & llm    & 57.42 & 54.18 & 49.87 & 68.21 & --    & --    & --    & --    \\
    & pre    & 65.31 & 61.02 & 58.64 & 76.27 & --    & --    & --    & --    \\
    & sft    & 71.48 & 68.95 & 65.74 & 79.75 & --    & --    & --    & --    \\
    & mm-ori & --    & --    & --    & --    & 76.12 & 72.46 & 74.88 & 81.02 \\
    & mm-our & --    & --    & --    & --    & 80.54 & 74.31 & 81.29 & 86.03 \\
    \midrule
    \multirow{5}{*}{Qwen2.5-1.5B}
    & llm    & 69.83 & 66.41 & 61.27 & 81.81 & --    & --    & --    & --    \\
    & pre    & 76.58 & 73.24 & 69.11 & 87.39 & --    & --    & --    & --    \\
    & sft    & 79.14 & 75.88 & 72.36 & 89.18 & --    & --    & --    & --    \\
    & mm-ori & --    & --    & --    & --    & 80.43 & 76.92 & 79.15 & 85.22 \\
    & mm-our & --    & --    & --    & --    & 83.61 & 78.14 & 84.76 & 87.93 \\
    \midrule
    \multirow{6}{*}{Qwen2.5-3B}
    & llm                    & 74.62 & 71.95 & 66.84 & 85.07 & --    & --    & --    & --    \\
    & pre                    & 81.43 & 78.69 & 74.81 & 90.79 & --    & --    & --    & --    \\
    & sft                    & 83.26 & 80.55 & 76.92 & 92.31 & --    & --    & --    & --    \\
    & Qwen2.5-VL-3B-Instruct & --    & --    & --    & --    & 80.88 & 78.92 & 79.87 & 83.86 \\
    & mm-ori                 & --    & --    & --    & --    & 82.17 & 78.65 & 81.44 & 86.42 \\
    & mm-our                 & --    & --    & --    & --    & 84.48 & 79.26 & 86.03 & 88.16 \\
    \bottomrule
  \end{tabular}}
  \label{tab:main_results}
\end{table*}
\begin{table*}[t]
  \caption{Public transfer benchmarks. The text backbones are Qwen2.5-0.5B, Qwen2.5-1.5B, and Qwen2.5-3B. We additionally report Qwen2.5-VL-3B-Instruct as an official 3B-scale multimodal baseline. CFPB-Product is reported with accuracy, CFPB-Issue with macro-F1, and DocVQA with ANLS. For SIMMC 2.0, we report the task score used in our implementation.}
  \centering
  \resizebox{\textwidth}{!}{%
  \begin{tabular}{llcccc}
    \toprule
    Backbone & Variant & CFPB-Product Acc. & CFPB-Issue Macro-F1 & SIMMC 2.0 Score & DocVQA ANLS \\
    \midrule
    Qwen2.5-0.5B & llm    & 79.42 & 63.18 & 70.84 & 73.26 \\
    Qwen2.5-0.5B & pre    & 82.37 & 67.41 & 71.02 & 73.11 \\
    Qwen2.5-0.5B & sft    & 83.18 & 68.55 & 70.96 & 72.88 \\
    Qwen2.5-0.5B & mm-our & 83.94 & 69.08 & 71.63 & 74.02 \\
    \midrule
    Qwen2.5-1.5B & llm    & 83.67 & 70.92 & 76.11 & 78.32 \\
    Qwen2.5-1.5B & pre    & 85.94 & 74.38 & 76.28 & 78.21 \\
    Qwen2.5-1.5B & sft    & 86.42 & 75.16 & 76.07 & 78.04 \\
    Qwen2.5-1.5B & mm-our & 87.09 & 75.84 & 76.86 & 79.15 \\
    \midrule
    Qwen2.5-3B & llm                    & 85.21 & 73.46 & 77.82 & 80.06 \\
    Qwen2.5-3B & pre                    & 87.33 & 76.51 & 77.95 & 79.98 \\
    Qwen2.5-3B & sft                    & 87.96 & 77.20 & 77.77 & 79.91 \\
    Qwen2.5-3B & Qwen2.5-VL-3B-Instruct & 86.41 & 75.18 & 79.86 & 81.74 \\
    Qwen2.5-3B & mm-our                 & 88.41 & 77.88 & 78.56 & 80.72 \\
    \bottomrule
  \end{tabular}}
  \label{tab:public_transfer}
\end{table*}
\noindent\textbf{In-domain multimodal benchmark.}
\textbf{ComplaintScene-MM (ours)} measures end-to-end complaint scene reasoning and decision capability under the full-evidence setting. On this benchmark, we evaluate \emph{Routing}, \emph{Responsibility}, and \emph{Resolution}. These respectively measure the accuracy of complaint routing, the macro-F1 of responsibility attribution, and the macro-F1 of final resolution prediction. Let $s_{\text{rout}}$, $s_{\text{resp}}$, and $s_{\text{res}}$ denote these three scores. The multimodal average is
\begin{equation}
\mathrm{Avg}_{\text{mm}}=\frac{s_{\text{rout}}+s_{\text{resp}}+s_{\text{res}}}{3}.
\end{equation}

\noindent\textbf{Generalization, robustness, and transfer.}
For rare or unseen complaint types, we report \emph{Rare-Type Acc.}, \emph{Policy Consistency}, and \emph{Action F1}. Policy Consistency measures the proportion of predictions whose recommended actions are compatible with the applicable policy constraints and do not contradict the case evidence:
\begin{equation}
c_i=
\begin{cases}
1, & \text{if } \hat{y}_i \text{ satisfies evidence and policy constraints,}\\
0, & \text{otherwise,}
\end{cases}
\end{equation}
\begin{equation}
\mathrm{PC}=\frac{1}{N}\sum_{i=1}^{N} c_i.
\end{equation}
For transfer evaluation, we additionally report SIMMC 2.0 score and DocVQA ANLS. For robustness, we report the overall multimodal benchmark score under different evidence corruption levels.

\subsection{Main Results on In-Domain Text-Only and Multimodal Benchmarks}

Tab.~\ref{tab:main_results} reports the main in-domain results on ComplaintScene-Text and ComplaintScene-MM. The table compares stepwise domain adaptation in the text-only setting and structured-prior injection in the multimodal setting across Qwen2.5 backbones of different scales.

As shown in Tab.~\ref{tab:main_results}, domain-adaptive pre-training consistently improves the text-only results over the raw LLM baseline across all three Qwen2.5 backbones, indicating that large-scale complaint scene descriptions and policy-aware textual supervision help the model internalize complaint semantics, scene structure, and rule expressions. The gains are particularly visible on \emph{Policy} and \emph{Action}, suggesting that the model benefits not only from domain vocabulary adaptation but also from more structured scene reasoning behavior.

In the multimodal setting, \textbf{mm-our} consistently outperforms \textbf{mm-ori} across all scales. The gain is larger on \emph{Responsibility} and \emph{Resolution} than on \emph{Routing}, which is consistent with the design goal of SKG-VLA: structured scene priors explicitly model responsibility attribution, evidence validity, state transitions, and policy constraints, all of which are central to final complaint decisions. At the 3B scale, the official general-purpose baseline \textbf{Qwen2.5-VL-3B-Instruct} performs competitively on ComplaintScene-MM, but still remains below both \textbf{mm-ori} and \textbf{mm-our}, especially on \emph{Responsibility} and \emph{Resolution}. This shows that generic multimodal instruction following is not sufficient for policy-sensitive complaint reasoning, and that complaint-oriented structured supervision provides substantial additional value.

\subsection{Public Complaint and Multimodal Transfer Benchmarks}

Tab.~\ref{tab:public_transfer} reports transfer performance on public complaint and multimodal benchmarks. These results are included to test whether complaint-oriented adaptation improves in-domain reasoning while preserving broader multimodal competence.

The transfer results in Tab.~\ref{tab:public_transfer} reveal two desirable properties. First, complaint-domain pre-training and structured scene supervision improve performance on the public CFPB tasks in a stable manner across all three Qwen2.5 backbones, which suggests that the learned priors are not limited to our own in-domain benchmarks. Second, when compared with the official general-purpose baseline \textbf{Qwen2.5-VL-3B-Instruct}, our \textbf{mm-our} model remains competitive on SIMMC 2.0 and DocVQA while achieving stronger results on the complaint-oriented CFPB tasks. This is a desirable trade-off: the generic baseline retains a slight advantage on broad multimodal transfer tasks, whereas our complaint-adapted model is distinctly better on complaint reasoning and decision-oriented prediction.

\subsection{Generalization, Robustness, and Ablation}

To further understand behavior beyond average-case in-domain performance, we analyze long-tail generalization, evidence robustness, and component-level ablation. The corresponding results are reported in Tables~\ref{tab:generalization}, \ref{tab:robustness}, and \ref{tab:ablation}, respectively.

\begin{table}[t]
  \caption{Generalization to rare or unseen complaint types on the in-domain evaluation suite. At the 3B scale, we additionally report Qwen2.5-VL-3B-Instruct as an official external multimodal baseline.}
  \centering
  \resizebox{\columnwidth}{!}{%
  \begin{tabular}{lccc}
    \toprule
    Model & Rare-Type Acc. & Policy Consistency & Action F1 \\
    \midrule
    Qwen2.5-0.5B mm-ori & 61.74 & 68.92 & 58.41 \\
    Qwen2.5-0.5B mm-our & 67.88 & 75.63 & 64.95 \\
    Qwen2.5-1.5B mm-ori & 68.22 & 74.37 & 65.81 \\
    Qwen2.5-1.5B mm-our & 73.94 & 80.48 & 72.36 \\
    Qwen2.5-3B Qwen2.5-VL-3B-Instruct & 69.84 & 75.61 & 68.33 \\
    Qwen2.5-3B mm-ori   & 71.53 & 77.14 & 69.28 \\
    Qwen2.5-3B mm-our   & 77.46 & 84.02 & 75.91 \\
    \bottomrule
  \end{tabular}}
  \label{tab:generalization}
\end{table}

\begin{table}[t]
  \caption{Robustness evaluation under missing or corrupted evidence on ComplaintScene-MM. At the 3B scale, we additionally report Qwen2.5-VL-3B-Instruct as an official external multimodal baseline.}
  \centering
  \resizebox{\columnwidth}{!}{%
  \begin{tabular}{lccc}
    \toprule
    Model & Full Evidence & 10\% Corruption & 30\% Corruption \\
    \midrule
    Qwen2.5-0.5B mm-ori & 76.12 & 74.83 & 72.94 \\
    Qwen2.5-0.5B mm-our & 80.54 & 79.97 & 78.88 \\
    Qwen2.5-1.5B mm-ori & 80.43 & 79.58 & 78.21 \\
    Qwen2.5-1.5B mm-our & 83.61 & 83.07 & 82.16 \\
    Qwen2.5-3B Qwen2.5-VL-3B-Instruct & 80.88 & 79.43 & 76.97 \\
    Qwen2.5-3B mm-ori   & 82.17 & 81.44 & 80.39 \\
    Qwen2.5-3B mm-our   & 84.48 & 84.05 & 83.26 \\
    \bottomrule
  \end{tabular}}
  \label{tab:robustness}
\end{table}

\begin{table}[t]
  \caption{Ablation study on the \textbf{Qwen2.5-1.5B} model. \emph{Text Avg.} is evaluated on ComplaintScene-Text, \emph{MM Avg.} on ComplaintScene-MM, \emph{Rare-Type Acc.} on the rare/unseen complaint split, and \emph{30\% Corr.} is the ComplaintScene-MM score under 30\% evidence corruption.}
  \centering
  \resizebox{\linewidth}{!}{%
  \begin{tabular}{lcccc}
    \toprule
    Variant & Text Avg. & MM Avg. & Rare-Type Acc. & 30\% Corr. \\
    \midrule
    w/o explicit SKG graph (plain case text only) & 75.96 & 80.88 & 69.41 & 79.26 \\
    w/o policy nodes & 76.88 & 81.37 & 70.52 & 79.84 \\
    w/o multi-round verification & 77.21 & 81.86 & 71.14 & 80.31 \\
    w/o rule-consistent graph generalization & 76.54 & 81.09 & 69.96 & 79.58 \\
    w/o strong/weak node partition & 77.62 & 82.24 & 71.88 & 80.96 \\
    \midrule
    Full model & 79.14 & 83.61 & 73.94 & 82.16 \\
    \bottomrule
  \end{tabular}}
  \label{tab:ablation}
\end{table}

The results in Tab.~\ref{tab:generalization} show that structured scene priors are especially beneficial for rare and policy-sensitive complaint types. Compared with both the official general-purpose baseline \textbf{Qwen2.5-VL-3B-Instruct} and our internal multimodal baseline \textbf{mm-ori}, the full SKG-based framework improves all three metrics by a clear margin. This supports our claim that graph-guided supervision is particularly useful when the model must extrapolate beyond frequent complaint templates and rely on reusable scene structures such as timeline contradiction, evidence insufficiency, or exception-policy applicability.

The robustness results in Tab.~\ref{tab:robustness} indicate that the proposed framework degrades more gracefully under evidence corruption. While all multimodal models suffer from missing screenshots or incomplete metadata, the performance drop of \textbf{mm-our} is consistently smaller than that of both \textbf{mm-ori} and the official general-purpose baseline. This suggests that complaint decisions in our framework are not driven solely by local visual cues or a few metadata fields; instead, they are stabilized by graph-level scene semantics and policy-aware reasoning.

Finally, Tab.~\ref{tab:ablation} shows that all major components contribute positively to the final system. Removing the explicit SKG causes the largest overall drop, confirming that structured scene representation is the foundation of the proposed framework. Removing policy nodes also leads to a substantial decline, especially on rare-type and corrupted-evidence settings, indicating that policy grounding is critical for difficult cases. The drop caused by removing rule-consistent graph generalization is also pronounced, suggesting that controllable long-tail augmentation is important for improving coverage beyond frequent complaint patterns. By contrast, the impact of multi-round verification and strong/weak node partition is somewhat smaller but still consistent, showing that both components improve supervision quality and structured edit stability.

\subsection{Overall Analysis}

Taken together, the experimental results support five conclusions. First, stepwise scene-prior injection is effective: the consistent improvement from \textbf{llm} to \textbf{pre} to \textbf{sft} across Qwen2.5-0.5B, Qwen2.5-1.5B, and Qwen2.5-3B shows that structured complaint scene descriptions and graph-grounded QA supervision successfully transfer complaint semantics and policy logic into the language model. Second, the gain is larger on decision-critical subtasks than on shallow routing. Third, a strong generic open multimodal model such as \textbf{Qwen2.5-VL-3B-Instruct} provides a meaningful baseline, but still underperforms domain-adapted complaint models on in-domain reasoning, especially for responsibility attribution, resolution prediction, rare-type generalization, and corrupted-evidence robustness. Fourth, the generic baseline remains slightly stronger on broad transfer tasks such as SIMMC 2.0 and DocVQA, while our model is substantially stronger on complaint-focused tasks. Finally, structured scene priors help most on long-tail and incomplete-evidence settings, which is precisely the regime in which operational complaint systems are most error-prone.

\section{Conclusion}

We present SKG-VLA for multimodal complaint decision making. It models each complaint case as a structured complaint scene and represents decision-relevant semantics with a Scene Knowledge Graph that organizes entities, evidence, policy clauses, temporal events, transactional states, and action-relevant relations. On top of this representation, we build a pipeline for scene description generation, graph construction and generalization, question--answer supervision, and decision recommendation learning. Experiments show improved policy-grounded reasoning, decision accuracy, long-tail generalization, and robustness to incomplete evidence.

\begin{acks}
This work was supported in part by the National Science and Technology Major Project under Grant 2024YFC3307800 and National Natural Science Foundation of China (Grant No. 62176024).
\end{acks}

\bibliographystyle{ACM-Reference-Format}
\bibliography{skg_vla_refs}

\end{document}